\documentclass[preprint,12pt]{elsarticle}

\usepackage{algorithm}
\usepackage{algorithmic}

\usepackage{amssymb}
\usepackage{amsmath}

\usepackage{soul}
\usepackage{bm}
\usepackage{diagbox}
\usepackage{multirow}
\usepackage[dvipsnames]{xcolor}
\usepackage{xcolor}
\usepackage{dsfont}
\usepackage{amsfonts} 
\usepackage{cleveref}

\journal{Computer Vision and Image Understanding}

\begin{document}

\begin{frontmatter}



\title{Rehearsal-free and Task-free Online Continual Learning\\ With Contrastive Prompt}


\author{Aopeng Wang, Ke Deng, Yongli Ren and Jun Luo} 

\affiliation{organization={RMIT University},
            addressline={}, 
            city={Melbourne},
            postcode={}, 
            state={},
            country={}}

\affiliation{organization={RMIT University},
            addressline={}, 
            city={Melbourne},
            postcode={}, 
            state={},
            country={}}

\affiliation{organization={RMIT University},
            addressline={}, 
            city={Melbourne},
            postcode={}, 
            state={},
            country={}}

\affiliation{organization={Machine Intelligence Center},
            addressline={}, 
            city={Hong Kong},
            postcode={}, 
            state={},
            country={}}

\begin{abstract}
The main challenge of continual learning is \textit{catastrophic forgetting}. Because of processing data in one pass, online continual learning (OCL) is one of the most difficult continual learning scenarios. To address catastrophic forgetting in OCL, some existing studies use a rehearsal buffer to store samples and replay them in the later learning process, other studies do not store samples but assume a sequence of learning tasks so that the task identities can be explored. However, storing samples may raise data security or privacy concerns and it is not always possible to identify the boundaries between learning tasks in one pass of data processing. It motivates us to investigate rehearsal-free and task-free OCL (F2OCL). By integrating prompt learning with an NCM classifier, this study has effectively tackled catastrophic forgetting without storing samples and without usage of task boundaries or identities. The extensive experimental results on two benchmarks have demonstrated the effectiveness of the proposed method. 
\end{abstract}

\begin{graphicalabstract}
\includegraphics{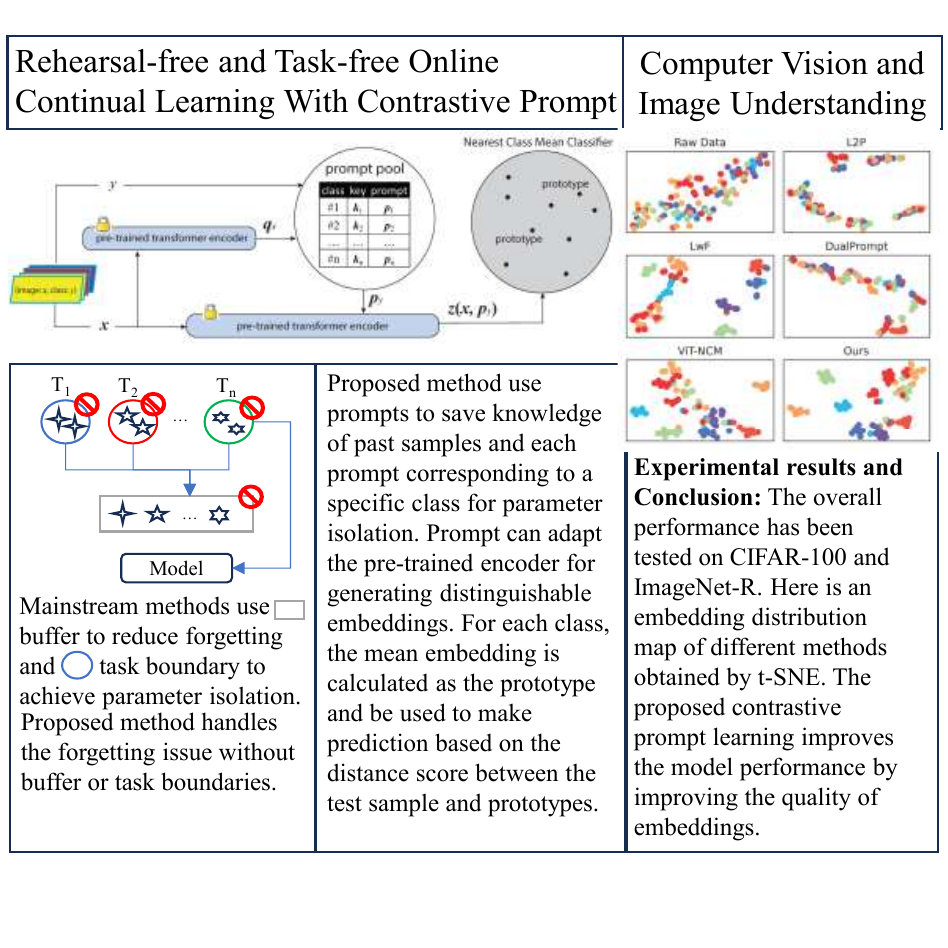}
\end{graphicalabstract}

\begin{highlights}
\item Integrates prompt learning with NCM classifier under rehearsal-free and task-free online continual learning.
\item Class-level prompts achieve parameter isolation and the contrastive loss guarantees the distinguishability between embeddings of different classes.
\item Evaluation in the CIFAR-100 and ImageNet-R datasets shows that the proposed method is effective for the rehearsal-free and task-free online continual learning scenario. 
\end{highlights}

\begin{keyword}
Online Continual Learning \sep Prompt Learning



\end{keyword}

\end{frontmatter}



\section{Introduction}
\label{sec:intro}
Continual learning (CL) \cite{de2021continual} has been widely investigated (see \cite{2023comprehensive} for a comprehensive survey). Continual learning is often implemented as a sequence of learning tasks. For example, the first task distinguishes whether an image is a cat or a dog, the next task distinguishes whether an image is a horse or a monkey, and so on. The main challenge is \textit{catastrophic forgetting} \cite{1989catastrophic}: the knowledge accumulated from previous learning tasks would become less informative as it continually integrates new knowledge from subsequent learning tasks. The most popular scenario is Class-incremental learning (CIL) \cite{2023comprehensive} where a single classifier is modeled for all tasks, and the task identity is available for training but not for inference \cite{2016progressive,2022cvote}. 

Recently, online continual learning (OCL) has attracted increasing attention from research communities. The rigorous OCL requires that the training data be processed in one pass and cannot be stored. However, most existing OCL studies assume a rehearsal buffer to store some informative samples called exemplars \cite{2020gdumb, de2021CoPE, 2019gradient, 2020dark, 2022info, 2022ocm, 2021scr, 2022cvt, 2022rar, 2022dvc, 2023onpro, 2022dsdm}. These rehearsal-based OCL studies are usually task-free (that is, they do not assume any information about task identities/boundaries) because it is often unrealistic in the online setting of continual learning to identify clear task boundaries in the training data processed in one pass \cite{2022dsdm}. 

It has been disclosed that storing samples in rehearsal buffer may cause privacy concerns \cite{2015privacy} and expensive memory cost \cite{2023summarizing, 2021memory}. The existing OCL methods that can work without rehearsal buffers are limited \cite{2022aop, 2020bld}. These rehearsal-free OCL studies make assumptions about task identities/boundaries as additional information to support effective solutions. In \cite{2022cvote}, the solution is based on storing samples in a rehearsal buffer and the task identifies. 
While it is plausible that OCL solutions should be (i) rehearsal-free to avoid privacy concerns of storing samples and (ii) task-free to circumvent the difficulty of identifying task boundaries in the training data processed in one pass, the rehearsal-free and task-free OCL solutions are less investigated.


This study proposes the rehearsal-free and task-free OCL (F2OCL) by integrating prompt learning and the nearest-class-mean (NCM) classifier. NCM classifier represents classes by the mean embeddings (prototype) of their samples. The NCM classifier and its variants have been widely used in few-shot or zero-shot learning \cite{10.1145/3386252,2018zero} and in continue learning \cite{2017icarl,de2021CoPE, 2021scr} where samples are stored in a memory buffer to address the catastrophic forgetting. 
Prompt learning is a popular technique stemming from natural language processing (NLP) \cite{2023pre}. Recently, prompt learning has been employed in off-line CIL \cite{Wang_2022_CVPR,2023decom,2023dual-pro,smith2023coda} where prompts are learnable parameters for tasks. Instead of storing samples, we propose to learn prompts for individual classes to address the catastrophic forgetting in the NCM classifier. The main contributions of this study are summarized as follows: 
\begin{enumerate}
  \item This study proposes to investigate task-free and rehearsal-free online continual learning (F2OCL), which avoids privacy concerns of storing samples and circumvents the difficulty of identifying task boundaries in training data processed in one pass. 
  \item This study introduces a framework to integrate prompt learning with the NCM classifier where prompts can be learned to address catastrophic forgetting without storing exemplars, and the NCM classifier enables us to continuously handle data without prior knowledge about task organization. 
  \item This study deliberately develops an algorithm to learn prompts so that the performance of the NCM classifier is optimized. The effectiveness has been demonstrated by extensive experiments.  
\end{enumerate}

\section{Related Works}
\label{sec:review}
Most methods of online continual learning are task-free but rehearsal-based\cite{2020gdumb, de2021CoPE, 2019gradient, 2020dark, 2022info, 2022ocm, 2021scr, 2022cvt, 2022rar, 2022dvc, 2023onpro, 2022dsdm}. They are task-free because it is unrealistic to identify clear task boundaries in the training data processed in one pass \cite{2022dsdm}. To mitigate catastrophic forgetting, they store samples from previous classes in a memory buffer and replay them in the learning process of new classes. One study \cite{2022info} devised strategies to select informative samples and prioritized the replay of informative samples while other studies \cite{de2021CoPE,2020gdumb} proposed to maintain the number balance of different class labels in the buffer. Besides, some methods \cite{2022dvc,2021scr,2022cvt} introduce contrastive learning to get better representations of images. In \cite{2017gradient,2018efficient}, they applied orthogonal projection so that the gradient directions for new classes are less parallel to those for previous classes when updating parameters to prevent performance drop on previous classes. Overall, rehearsal methods usually present state-of-the-art performance due to the stored samples from previous classes. But storing samples in rehearsal buffers may cause privacy concerns \cite{2015privacy} and expensive memory cost \cite{2023summarizing, 2021memory}. 

A few methods of OCL are rehearsal-free but task-based \cite{2020bld, 2022aop}. In \cite{2020bld}, Enrico et al. proposed Batch-level Distillation (BLD) to solve OCL under extreme memory constraints. For each task, BLD trains a specific classifier only for that task and executes a knowledge distillation between classifiers of the current task and the previous tasks to defy forgetting. To learn task-specific classifiers, they assume that the task boundaries between samples are known. In \cite{2022aop}, the proposed Adaptive Orthogonal Projection (AOP) learns for each task by updating the network parameters or weights only in the direction orthogonal to the subspace spanned by all previous tasks. In \cite{2022cvote}, the solution is based on storing samples in a rehearsal buffer and the task identifies. A feature buffer is built to store features, task identity, and class labels. The features are the embeddings of samples rather than the raw data. When learning for a new task, it will replay the sample embeddings of previous classes to mitigate forgetting for its classifier. The stored features and task identity are used to train a neural network to predict task identity for new samples. 

While it is plausible that OCL solutions should be (i) rehearsal-free to avoid privacy concerns of storing samples and (ii) task-free to circumvent the difficulty of identifying task boundaries in the training data processed in one pass, the rehearsal-free and task-free OCL solutions are less investigated. 

\section{Problem Statement}\label{sec:Preliminary}
The Rehearsal-free and Task-free Online Continual learning (F2OCL) aims at training a model on a sequence of sample batches \(\mathcal{B} = \{\mathcal{B}_1,\mathcal{B}_2,\cdots\}\). For the $t$-th batch, \(\mathcal{B}_t\) consists of a sequence of samples \({(\bm{x}_{i,t},y_{i,t})}^{N}_{i=1}\) where the number of samples in \(\mathcal{B}_t\) is $N$, \(\bm{x}_{i,t}\) is the image, and \(y_{i,t}\) is the label of \(\bm{x}_{i,t}\). For each label $y$, $[t_{ys}, t_{ye}]$ indicates that samples of label $y$ appear only from batch $\mathcal{B}_{t_{ys}}$ to $\mathcal{B}_{t_{ye}}$. The samples in each batch are processed in one pass and no samples from the previous batches are stored in memory, i.e., rehearsal-free. F2OCL model \(f_\theta\ : \mathcal{X} \rightarrow \mathcal{Y}\) predicts the label $y$ for an unseen image $\bm{x}$, i.e., \(y=f_\theta\ (\bm{x})\).  

\begin{figure*}[t]
  \centering
   \includegraphics[width=0.75\linewidth]{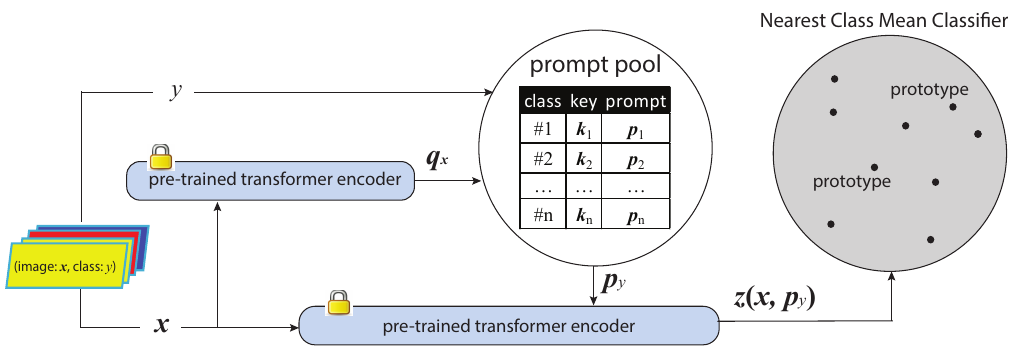}
   \caption{The proposed solution for F2OCL.}
   \label{fig2}
\end{figure*}

\section{Background}
\subsubsection*{Nearest Class Mean (NCM) Classifier}
The NCM classifier computes a class mean (prototype) vector for each class using the embeddings of samples in the class. 
\begin{equation}
    \bm{\mu}_c = \frac{1}{n_c}\sum_i f(\bm{x}_i)\mathds{1}\{y_i=c\}
\end{equation}
where $\bm{\mu}_c$ is the prototype of class $c$, $n_c$ is the number of samples for class $c$, and $\mathds{1}\{y_i = c\}$ is 1 if $y_i=c$, 0 otherwise. To predict a label for a new sample $\bm{x}$, NCM compares the embedding of $\bm{x}$ with all the prototypes and assigns $\bm{x}$ the class label of the 
nearest prototype: 
\begin{equation}
    y^* = \arg\min_{c\in C} \mathcal{D}(f(\bm{x}),\bm{\mu}_c)
\end{equation} 
where $C$ refers to the set of classes, $\mathcal{D}()$ is the metric function (usually L2 distance) and $f(\bm{x})$ is the embedding of $\bm{x}$. 

In \cite{2017icarl}, Rebuffi et al. proposed to approximate the prototype vectors using a convolutional neural network. In \cite{2021scr}, Mai et al. proposed contrast learning to optimize the encoder so that the embeddings of samples of the same class are similar and those of different classes are dissimilar. In \cite{de2021CoPE}, the prototype vectors incrementally evolve when new samples are available with contrastive learning used in a similary way. To address catastrophic forgetting issue, the methods in \cite{2021scr, de2021CoPE, 2017icarl} require some samples of all previous classes stored in a memory buffer. In \cite{2020semantic}, Yu et al. proposed a drift compensation to update previously computed prototypes of NCM classifier without using a memory buffer but it is task-based.

\subsubsection*{Prompt-based Continual Learning}
Prompt-based Continual Learning (Pb-CL) makes use of a rising technique called ``prompt learning" originating from NLP \cite{2023pre}, which is a parameter-efficient technique to adapt a pre-trained model to different downstream tasks. Recent studies \cite{Wang_2022_CVPR, 10.1007/978-3-031-19809-0_36, smith2023coda, 2023decom, 2024consistent, 2024rep} have applied prompt learning to address the forgetting problem in CIL by saving knowledge learned in previous tasks with prompts, rather than saving samples. Learning to Prompt (L2P) \cite{Wang_2022_CVPR} constructs a set of learnable prompts and selects $k$ prompts that are most similar to the new sample; then feed them with the sample to a pre-trained model (a vision transformer) to enhance the sample representation to eventually improve prediction accuracy. Dual-P \cite{10.1007/978-3-031-19809-0_36} designed two kinds of prompt, \textit{G(eneral)-Prompt} and \textit{E(xpert) prompt}, to encode task-invariant and task-specific instructions, from the perspective of Complementary Learning Systems \cite{1995cls,2016cls}. CODA-P \cite{smith2023coda} modified the prompt selection, from selecting the top-$k$ most similar ones to setting an adaptive weight vector for all prompts. \cite{2024rep} and \cite{2025pc} also focused on improving the prompt selection part, the former proposed introduces adaptive token merging (AToM) and layer dropping (ALD) to minimize computational costs at inference, and the latter proposed a customization process for each instance to generate adaptive prompts, avoiding the hard selection of prompts. HiDe-Prompt\cite{2023decom} further decomposes the CIL problem into hierarchical components: within-task prediction, task-identity inference, and task-adaptive prediction. Then it solves three components jointly. However, these methods are designed for offline Class-incremental continual learning, which is task-based and cannot solve the F2OCL problem by default. 

\section{Methodology}
The framework of the proposed solution for F2OCL is presented in Figure \ref{fig2}. It consists of three components, a pre-trained vision transformer (ViT) encoder, a prompt pool, and a nearest-class-mean classifier. The pre-trained transformer encoder is used to produce the generalizable representations following previous work \cite{10.1007/978-3-031-19809-0_36, Wang_2022_CVPR}. To maintain its generality, the pre-trained transformer encoder is frozen, i.e., its parameters are not updated. Figure \ref{fig2} presents two pre-trained transformer encoders that are the same. One produces the key for the image $\bm{x}$ and the other outputs embedding of the image $\bm{x}$ augmented with prompt $\bm{p_y}$ where $y$ is the label of $\bm{x}$. The prompt pool contains a set of triplets in the form of $(class, key, prompt)$ where each $key$ is a vector $\bm{k}\in \mathbb{R}^{d}$ and each $prompt$ is a vector $\bm{p}\in \mathbb{R}^{L_p\times d}$, with $L_p$ representing the length of a prompt. 


\subsection{F2OCL Training}
Each prompt in the prompt pool corresponds to one class, and vice versa. Initially, the prompt pool is empty. The training samples are processed in batches. When processing the current batch $\mathcal{B}=(\bm{x}_i,y_i)_{i=1}^{n}$, if a sample comes with a new class $y$, a triplet $(\text{class:}y, \text{key:}\bm{k}_{y}, \text{prompt:}\bm{p}_{y})$ is created and inserted in the prompt pool where $\bm{k}_y$ and $\bm{p}_y$ are initialized, and a new prototype in NCM classifier is created and initialized; if a sample comes with an existing class $y$, the corresponding triplet $(\text{class:}y, \text{key:}\bm{k}_{y}, \text{prompt:}\bm{p}_{y})$ is optimized, and the relevant prototypes in the NCM classifier are optimized as well. The pseudo-code for F2OCL training is presented in Algorithm \ref{alg:ContrastPrompt} 

\begin{algorithm}
\caption{F2OCL Training}
\label{alg:ContrastPrompt}
\begin{algorithmic}[1]
\STATE \textbf{Input:} A pre-trained vision transformer encoder $f(.)$, the training batches $\mathcal{B}=\{\mathcal{B}^1,\mathcal{B}^2,\cdots\}$, a prompt pool $\mathcal{P}$, a set of prototypes $\mathcal{M}$ in the NCM classifier.
\STATE \textbf{Initialize:} $\mathcal{P}=\emptyset$, $\mathcal{M}=\emptyset$   
\FOR{$t$ in range(len($\mathcal{B}$))}
\STATE Load batch $B^t$ in memory 
\FOR{each sample $(\bm{x}_i,y_i)\in B^t$}
 \IF{$y_i$ is a new class in $\mathcal{P}$}
   \STATE Create and initialize $\bm{k}_{y_i}$ and $\bm{p}_{y_i}$
   \STATE Create a new prototype $\mu_{y_i}= f(\bm{x}_i,\bm{p}_{y_i})$
   \STATE Insert $(y_i,\bm{k}_{y_i},\bm{p}_{y_i})$ to $\mathcal{P}$
   \STATE Insert $\mu_{y_i}$ to $\mathcal{M}$
 \ENDIF
 \STATE $z_{\bm{x}_i} = f(\bm{x}_i,\bm{p}_{y_i})$
 \STATE Update prompts using loss Equation \ref{cploss}
 \STATE Update keys using Equation \ref{keyupdate}
\ENDFOR
\STATE Update NCM classifier using Equation \ref{meanembedding}
\ENDFOR
\end{algorithmic}
\end{algorithm}

\subsubsection*{Prompt Contrastive Learning}
Each sample $(\bm{x}_i,y_i)$ in the current batch $\mathcal{B}$ is processed in one pass. 
Let $\bm{z}_{x_i}=f(\bm{x}_i,\bm{p}_{y_i})$ represent the embedding of image $\bm{x}_i$ augmented with the corresponding prompt $\bm{p}_{y_i}$. For the sake of presentation, we simply call $\bm{z}_{x_i}$ the augmented embedding of $\bm{x}_i$. We use $\bm{\mu}_{y_i}$ to denote the prototype in the NCM classifier, i.e., the center of augmented embeddings for all samples of the class $y_i$ encountered in the batches processed so far.

Intuitively, for the samples of the same class, their augmented embeddings should be close to each other; for the samples of the different classes, their augmented embeddings should be far away from each other. That is, for different classes, it is desired that the centers of augmented embeddings are as dispersed as possible. The situation is complex here. Because batches are processed continuously, we need to consider samples across different batches rather than the current batch only. In F2OCL, batches are processed in one pass and no samples in previous batches are stored. What is retained are the centers of the augmented embeddings by processing the previous batch(es), i.e., a set of centers $\{\bm{\mu}_{y_1},\bm{\mu}_{y_2},\cdots\}$, each for one class encountered so far. When processing the samples in the current batch, we should consider not only their augmented embeddings but also the centers of augmented embeddings from the previous batch(es), if they exist. 


To effectively learn prompts, we introduce prompt contrastive learning. For each sample $(\bm{x}_i,y_i)$ in the current batch $\mathcal{B}$, we construct the positive set $S^+_{\bm{x}_i}$ and negative set $S^-_{\bm{x}_i}$. All samples in $\mathcal{B}$ whose class is $y_i$ forms $S^+_{\bm{x}_i}$ and other samples constitute $S^-_{\bm{x}_i}$. For samples 
in the current batch $\mathcal{B}$, the loss function for prompt contrastive learning is below:
\begin{align}\label{cploss}
L_{cp} & = -(\alpha\textstyle\log \Lambda_1+\frac{ \beta}{\left | S^+_{\bm{x}_i} \right | }\sum_{(\bm{x}_j,y_j)\in S^+_{\bm{x}_i}}\textstyle\log \Lambda_2),\\
\Lambda_1 & = \frac{\exp (\delta(\bm{z}_{\bm{x}_i},\bm{\mu}_{y_i}))}{\exp (\delta(\bm{z}_{\bm{x}_i},\bm{\mu}_{y_i}))+\Gamma_1},\\
\Gamma_1 & = \textstyle\sum_{(\bm{x}_j,y_j)\in S^-_{\bm{x}_i}}\exp (\delta(\bm{z}_{\bm{x}_i},\bm{\mu}_{y_j})),\\
\Lambda_2 & = \frac{\exp (\delta(\bm{z}_{\bm{x}_i},\bm{z}_{\bm{x}_j}))}{\sum_{(\bm{x}_j,y_j)\in S^+_{\bm{x}_i}}\exp (\delta(\bm{z}_{\bm{x}_i},\bm{z}_{\bm{x}_j}))+\Gamma_2},\\
\Gamma_2 & = \textstyle\sum_{(\bm{x}_k,y_k)\in S^-_{\bm{x}_i}}\exp (\delta(\bm{z}_{\bm{x}_i},\bm{z}_{\bm{x}_k})).
\end{align}
where $\bm{z}_{\bm{x}} = f(\bm{x},\bm{p}_{y})$ is the augmented embedding of $\bm{x}$, $\bm{\mu}_{y}$ is the prototype of class $y$, $\left | S^+_{\bm{x}_i} \right |$ is the number of samples in $S^+_{\bm{x}_i}$, and cosine similarity $\delta(.,.)$ is used to measure similarity between embeddings. The loss does not directly optimize the prototypes of classes in the NCM classifier, but it optimizes the prompts so that the augmented embeddings of samples are updated to disperse the prototypes of different classes.  

\begin{figure}[t]
  \centering
   \includegraphics[width=1\linewidth]{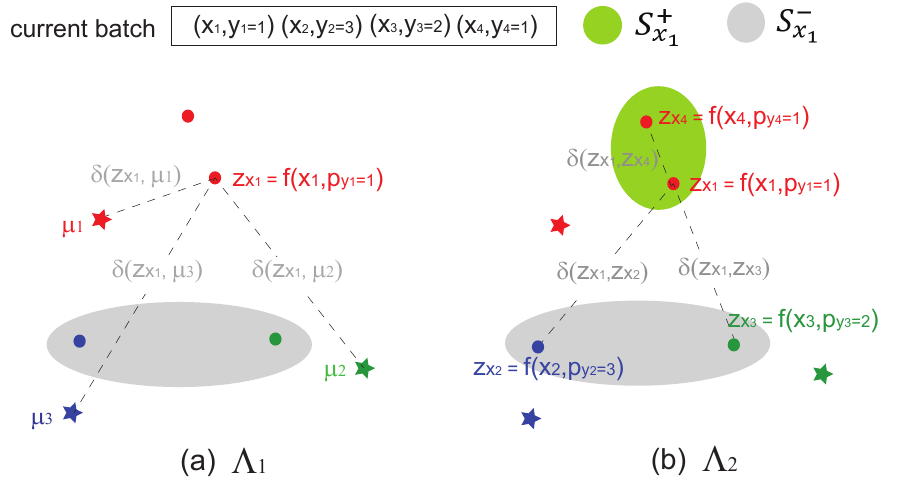}
   \caption{Contrastive prompt learning.}
   \label{fig4}
\end{figure}

$\Lambda_2$ is contrastive learning loss for sample $(\bm{x}_i,y_i)$ in the current batch $B$. As illustrated by the example in Figure \ref{fig4}(b), the augmented embedding $\bm{z}_{\bm{x}_1}$ of a sample $(\bm{x}_1,y_1)$ in the current batch should be closer to $\bm{z}_{\bm{x}_4}$ due to the same label and far away from $\bm{z}_{\bm{x}_2}$ and $\bm{z}_{\bm{x}_3}$ due to different label. $\Lambda_1$ is loss for sample $(\bm{x}_i,y_i)$ based on the prototypes in the NCM classifier. In Figure \ref{fig4}(a), $\bm{z}_{\bm{x}_1}$ should be closer to the prototype of the same label $\bm{\mu}_1$ and far away from other prototypes. The prototype represents the running mean embeddings of samples of the same label in previous batches. For $(\bm{x}_i,y_i)$, if the prototype $\bm{\mu}_{y_i}$ represents more samples, more weight $\alpha$ is allocated to $\Lambda_1$; otherwise, less weight. In out study, $\alpha=\frac{n_{y_i}}{n_{y_i}+n^*_{y_i}}$ where $n_{y_i}$ is the number of samples of class $y_i$ in the previous batch(es), $n^*_{y_i}$ is the number of samples of $y_i$ in the current batch $B$. For each sample of class $y_i$ in the current batch $B$, the weight $\beta=\frac{n^*_{y_i}}{n_{y_i}+n^*_{y_i}}$ is allocated.  
\subsubsection*{Update NCM Classifier}
If the prompt $\bm{p}_{y_i}$ in the prompt pool has been updated based on samples in the current batch $\mathcal{B}$, the prototype $\bm{\mu}_{y_i}$ in the NCM Classifier is updated accordingly. Let $\bm{p}'_{y_i}$ be the updated prompt after processing batch $\mathcal{B}$. For each sample $\bm{x}_i$ of label $y_i$ in batch $B$, the augmented embedding $\bm{z}'_{x_i}=f({\bm{x}_i},\bm{p}'_{y_i})$ is created based on the updated prompt $\bm{p}'_{y_i}$. Let $\bm{\mu}'_{y_i}$ be the updated prototype. We have:
\begin{equation}\label{meanembedding}
\bm{\mu}'_{y_i} = \frac{n_{y_i}\bm{\mu}_{y_i} + \sum_{(\bm{x},y=y_i)\in B}\bm{z}'_{\bm{x}}}{n_{y_i}+n^*_{y_i}}.  
\end{equation} 

\subsubsection*{Key Update}
In the prompt pool, the key of a triplet represents all samples with the same label across batches processed so far. When a new class label $y_i$ is encountered in the current batch $B$, a new triplet $(y_i,\bm{k}_{y_i},\bm{p}_{y_i})$ is created. The key $\bm{k}_{y_i}$ is randomly initialized. 

If the triplet already exists $(y_i,\bm{k}_{y_i},\bm{p}_{y_i})$, key $\bm{k}_{y_i}$ is updated based on the samples labeled $y_i$ in the current batch $B$ by minimizing loss $L_{key}$:
\begin{equation}\label{keyupdate}
    L_{key} = \alpha\delta(\bm{k}'_{y_i}, \bm{k}_{y_i})+\beta\sum_{(\bm{x}_i,y_i)\in B}\delta(\bm{k}'_{\bm{y}_i}, \bm{q}_{\bm{x}_i}).
\end{equation}
where $\mathbf{k}_{y_i}$ is the key before processing batch $\mathcal{B}$, $\bm{k}'_{y_i}$ is initialized as $\mathbf{k}_{y_i}$ and updated to minimize $L_{key}$ 
by performing the gradient descent iteratively. Since samples in a batch are processed in one pass, $\bm{k}'_{y_i}$ is updated once only\footnote{The second term of $L_{key}$ works only in this situation.}. The updated $\bm{k}'_{y_i}$ is the key after processing the current batch $\mathcal{B}$.   

\subsection{Classification with Prompt-based NCM Classifier}
Given a new sample $\bm{x}$, the pre-trained vision transformer (ViT) encoder $f(.)$ is used to generate $\bm{q}_{\bm{x}} = f(\bm{x})$. The most similar key from the prompt pool is retrieved.
\begin{equation}
    \bm{k} = \arg\min_{\bm{k}_i\in \mathcal{P}}\delta(\bm{q}_{\bm{x}},\bm{k}_i).
\end{equation}
Sample $\bm{x}$ and the prompt $\bm{p}$ associated with key $\bm{k}$ are used to generate the augmented image embedding $\bm{z}_{\bm{x}}=f(\bm{x},\bm{p})$. In the NCM classifier, the closest prototype to $\bm{z}_{\bm{x}}$ is the predicted class of $\bm{x}$.
\begin{equation}\label{eq:11}
    y = \arg\min_{y_i}\delta(\bm{\mu}_{y_i},\bm{z_x}).
\end{equation}
To improve the performance, 
one may consider to select $\mathcal{K}$ most similar keys of $\bm{q}_{\bm{x}}$ from the prompt pool, and then concatenate the corresponding $\mathcal{K}$ prompts into a long prompt as in \cite{Wang_2022_CVPR}. With sample $\bm{x}$ and the concatenated prompt as input, the pre-trained vision transformer (ViT) encoder $f(.)$ generates an augmented embedding, which is used to search for the nearest prototype in the NCM classifier as the predicted class of $\bm{x}$. 

\section{Experiments}
All experiments are implemented using Pytorch \cite{2019pytorch} and conducted on a single Tesla V100. The code will be available upon acceptance.  

\begin{table*}[htb]
\centering
\small
\begin{tabular}{l|l|l|ll|ll}
\hline
\multirow{2}{*}{PTM} & \multirow{2}{*}{Method} & \multirow{2}{*}{Buffer Size} & \multicolumn{2}{l|}{CIFAR-100} & \multicolumn{2}{l}{ImageNet-R} \\
                         &                  &   & \(A_n\)(↑)  & \(F_n\)(↓) & \(A_n\)(↑) & \(F_n\)(↓) \\ \hline
\multirow{6}{*}{Sup-21K} & ER               & 200 & 53.63±0.87 & 45.20±1.03 & 44.91±0.67 & 42.83±0.57 \\
                         & LwF              & 0 & 9.54±1.37 & 13.76±1.95 & 7.04±0.57 & 11.53±1.70    \\
                         & DualPrompt       & 0 & 30.05±1.98 & 72.12±2.25 & 14.67±0.24 & 50.18±0.55 \\
                         & L2P              & 0 & 29.83±1.65 & 68.90±2.29 & 22.85±0.12 & 38.87±0.48 \\
                         & ViT-NCM          & 0 & 66.82±0.17 & 9.35±0.40 & 45.44±0.22 & \textbf{7.50}±0.53 \\
                         & Ours             & 0 & \textbf{71.18}±0.18 & \textbf{8.22}±0.55 & \textbf{46.75}±0.14 & 7.80±0.47  \\ \hline
\multirow{6}{*}{DINO-1K} & ER               & 500 & 48.66±0.74 & 43.66±1.39 & 39.48±3.22 & 38.92±3.51  \\
                         & LwF              & 0 & 7.99±2.26 & \textbf{11.46}±2.66 & 5.65±0.73 & 12.43±1.69 \\
                         & DualPrompt       & 0 & 14.83±0.65 & 86.80±0.22 & 11.80±0.67 & 73.46±1.23 \\
                         & L2P              & 0 & 11.39±0.20 & 88.17±0.38 & 10.31±0.13 & 71.64±0.70 \\
                         & ViT-NCM          & 0 & 51.91±0.11 & 12.56±0.58 & 43.63±0.05 & 9.08±0.21  \\
                         & Ours             & 0 & \textbf{55.26}±0.60 & 12.02±0.94 & \textbf{44.11}±0.30 & \textbf{7.72}±0.24  \\ \hline
\end{tabular}
\caption{Overall performance measured by the average accuracy (\(A_n\)) and average forgetting \(F_n\) (with ± standard deviation over three runs of different random seeds and group splits).}

\label{tab:overall}
\end{table*}

\begin{table}[htb]
\centering
\footnotesize 
\begin{tabular}{|c|ccc|ccc|}
\hline
\multirow{2}{*}{PTM} & \multicolumn{3}{c|}{CIFAR-100} & \multicolumn{3}{c|}{ImageNet-R} \\ \cline{2-7} 
        & \multicolumn{1}{c|}{\(A_k\)(↑)} & \multicolumn{1}{c|}{\(A_n\)(↑)}  & UB    & \multicolumn{1}{c|}{\(A_k\)(↑)} & \multicolumn{1}{c|}{\(A_n\)(↑)}    & UB      \\ \hline
Sup-21K & \multicolumn{1}{c|}{0.77}     & \multicolumn{1}{c|}{70.0}  & 71.43 & \multicolumn{1}{c|}{0.52}   & \multicolumn{1}{c|}{46.18}   & 46.56   \\ \hline
DINO-1K & \multicolumn{1}{c|}{1.04}     & \multicolumn{1}{c|}{52.12} & 53.89 & \multicolumn{1}{c|}{0.45}   & \multicolumn{1}{c|}{43.93} & 44.93 \\ \hline
\end{tabular}
\caption{Impact of key selection accuracy.}
\label{tab:key_impact}
\vspace{-0.3cm}
\end{table}

\textbf{Datasets} The experiments are conducted on two benchmark datasets, CIFAR-100 \cite{cifar100} and ImageNet-R \cite{imagenet}, widely used in continual learning studies. CIFAR-100 includes 100 classes of small-scale images, with 500 training samples and 100 test samples for each class. We organize training samples into a sequence of batches $\mathcal{B}=\{\mathcal{B}_1,\mathcal{B}_2,\cdots,...\mathcal{B}_n\}$, where the batch size is 10 following the general setting in online continual learning studies, and thus $n=5000$. The first $500$ batches, i.e., the first group of classes, include the samples of 10 randomly selected classes, the second $500$ batches, i.e., the second group of batches, are samples of another 10 different classes selected randomly, and so on. In total, we have 10 groups of batches and no samples in different groups have the same class. 

ImageNet-R consists of 30000 large-scale images of 200 classes that are hard examples of ImageNet \cite{imagenet} or newly collected examples of different styles. Different classes have different numbers of samples. For each class, $80\%$ of samples are used as training data and $20\%$ as test data. We randomly select 20 classes and organize the samples of these classes in training data into $m_1$ batches, i.e., the first group of batches, where the batch size is 10 and $m_1$ is determined by the number of samples. Then, 20 different classes are randomly selected and their samples are organized into $m_2$ batches, i.e., the second group of batches. This process repeats until all classes have been selected and their samples are organized into batches. In total, we have 10 groups of batches and no samples in different groups have the same class. The 10 groups of batches are randomly concatenated in sequence.   


\textbf{Evaluation metrics} We apply two evaluation metrics widely used in continual learning studies, e.g., \cite{Wang_2022_CVPR,10.1007/978-3-031-19809-0_36}. The two metrics are defined as follows:
\begin{align}
\label{metrics}
    A_n&=\frac{1}{n}\sum_{\tau=1}^{n}\mathcal{T}_{n,\tau} \\
    F_n&=\frac{1}{n-1}\sum_{\tau=1}^{n-1}\max_{{\tau}'\in{1,\cdots,n-1}}(\mathcal{T}_{{\tau}',\tau }-\mathcal{T}_{n,\tau})  
\end{align}

The first metric is \textit{Average Final Accuracy} (\(A_n\)) and the second metric is \textit{Average Forgetting} (\(F_n\)). After the model is trained using training samples in the first $n$ groups of batches, \(A_n\) is the average classification accuracy across 
the first $n$ groups of batches. 
\(\mathcal{T}_{n,\tau}\) is the classification accuracy across the test samples whose classes appear in the $\tau$-th group of batches. 
\(A_n\) is the average of \(\{\mathcal{T}_{n,1}, \cdots, \mathcal{T}_{n, n}\}\). 
When the model is incrementally trained using more training samples of new classes, the classification accuracy for the test samples of old classes tends to drop, i.e., catastrophic forgetting. \(F_n\) measures the average classification accuracy drop when the model is incrementally trained from the first group to $n$-th group. The higher $A_n$ and the lower \(F_n\) indicate the better performance. 

\textbf{Baselines} We compare the performance of the proposed prompt-based NCM classifier (denoted as Ours) with baselines including the state-of-the-art solutions. 
\begin{enumerate}
    \item \textbf{Regularization-based method:} Learning without Forgetting (LwF) \cite{2017lwf} is a widely compared method, which uses the knowledge distillation technique to preserve the knowledge from past samples. 
    \item \textbf{Rehearsal-based method:} Experience Replay (ER) \cite{2019er} is a simple yet effective baseline, which trains the model with a combination of the current input batch and an extra batch from the memory buffer that stores history samples. 
    \item \textbf{Prompt-based methods:} Learning to Prompt (L2P) \cite{Wang_2022_CVPR} and DualPrompt \cite{10.1007/978-3-031-19809-0_36} are the first two studies that bring prompt learning into class incremental continual learning. L2P constructs a (\textit{key, prompt}) pair pool and learns the knowledge for each learning task with a combination of prompts. DualPrompt further divides the prompts into a general prompt and expert prompts for learning task-invariant knowledge and task-specific knowledge, respectively.
    \item \textbf{ViT-NCM:} We 
    build up an NCM classifier by embedding samples using the pre-trained ViT encoder and using their prototypes. No prompts are involved.  
\end{enumerate}

Following the previous studies \cite{Wang_2022_CVPR,10.1007/978-3-031-19809-0_36}, we adopt a pre-trained ViT-B/16 backbone and train with an Adam optimizer (\(\beta_1\) = 0.9, \(\beta_2\) = 0.999), a batch size of 10, 
a constant learning rate of 0.1, prompt length of 20, temperature parameter of 0.2. The images are resized to 224 × 224 and normalized to [0, 1]. For a fair comparison, all baselines use the same pre-trained ViT encoder that is frozen. For all baselines, we carefully follow the best parameter settings in the original papers, if they exist. Data is processed in one pass for all methods during training. Our method selects one key from the prompt pool during testing and for other prompt-based methods, the number of keys selected follows the best setting of the original papers.

\subsection{Overall Performance}
The \cref{tab:overall} shows the performances of all methods on two datasets and two different pre-trained weights: Sup-21K\cite{2021imagenet} and DINO-1K\cite{2021dino}. Sup-21K is the weights trained on ImageNet-21K in a supervised way, and DINO-1K is the weights trained on ImageNet-1K in an unsupervised way. 

Compared to the methods based on linear classifiers, the methods based on the NCM classifier (ViT-NCM and Ours) usually have better performances on both datasets, regarding average accuracy and average forgetting. The results demonstrate that the linear classifier is more easily biased towards the recent samples, i.e., forgetting, than the NCM classifier. Since this study concerns the rehearsal-free and task-free OCL, it is unfair to our method to compare with ER which depends on rehearsal buffer. However, ER is compared to show whether the proposed prompts can capture information as a rehearsal buffer. 

When the pre-trained weights are changed from Sup-21K to DINO-1K, all methods present different levels of performance drop on both datasets. Among them, ER and LwF present a smaller drop. For ER, the buffer size has enlarged from 200 to 500 to reduce performance drop. For LwF, its performance under Sup-21K is already awful and the space to further drop is limited. 
In conclusion, the pre-trained weights are crucial to the performance. 

By default, our method selects one key, i.e., one prompt, from the prompt pool during testing. This setting is based on empirical studies reported later in subsection \textit{multiple passes and keys}. One may ask whether the accuracy of key selection determines the performance of our method. In \cref{tab:key_impact}, the accuracy of key selections $A_k$ is reported as very low. One reason is that data is processed in one pass during training and the other reason is that the number of keys is large, one key for each class. In contrast, the average accuracy $A_n$ is 40-70 times higher. It is reasonable to attribute the significant improvement to prompts. The column \text{UB} is the average accuracy $A_n$ when the key selection is $100\%$ correct. Interestingly, \text{UB} is marginally better than $A_n$. This further evidences that the accuracy of key selection has a trivial impact on the performance, but is determined by the prompts learned. 

\begin{figure}[htb]
    \centering
    \includegraphics[width=1\linewidth]{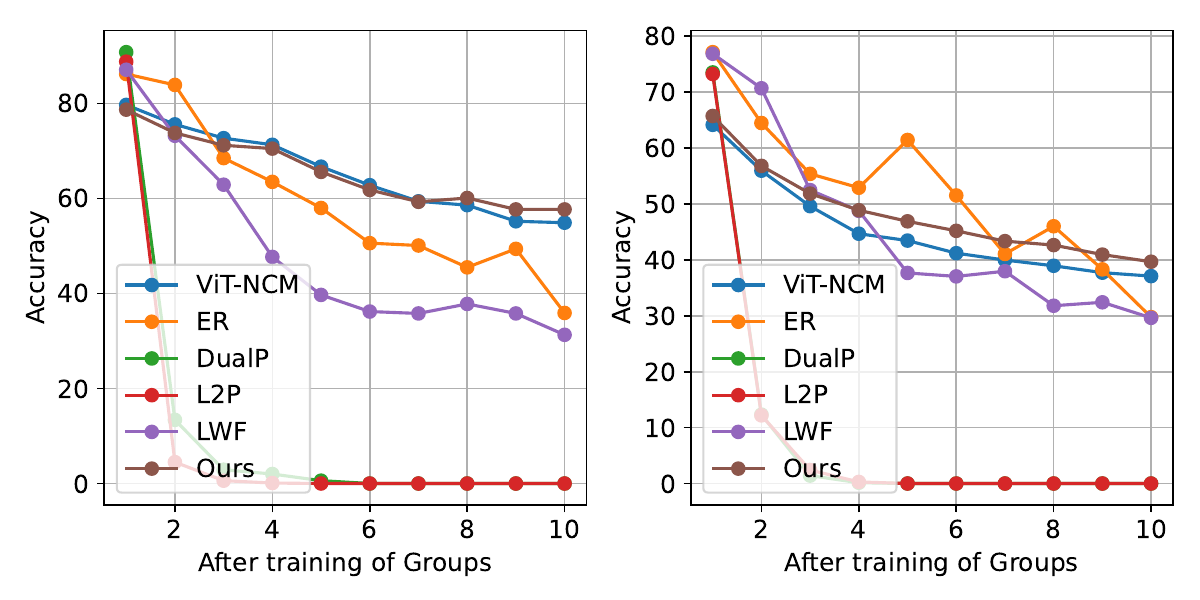}
    \caption{Forgetting resistance on CIFAR-100 (left) and on ImageNet-R (right) with weights DINO-1K.}
    \label{forget}
    \vspace{-0.4cm}
\end{figure}

\subsection{Forgetting Resistance}
To better understand the performance of different methods in terms of forgetting, we train our model and baselines incrementally using more groups of batches in training data and test the samples of the classes only in the first group of batches. 
The test results are reported in \cref{forget}. After being trained by one more group of batches, all methods demonstrate forgetting, i.e., accuracy \(\mathcal{T}_{x,1}\) decreases. Our method demonstrates the best performance in forgetting resistance, i.e., the slowest accuracy decrease. For ER on ImageNet-R, it has a comparable performance as our method in \cref{forget}. However, if looking at the overall performance on $F_n$ in \cref{tab:overall}, our method beats ER on ImageNet-R. We investigate why ER on ImageNet-R demonstrates a better performance than on CIFAR-100. Compared with CIFAR-100, ImageNet-R consists of hard samples. The hard samples stored in the rehearsal buffer of ER provide information that is hard to capture by the prompts in our method. This may be the reason.   

\subsection{Effect of Prompt}
Compared with other prompt-based methods L2P and DualPrompt, our method demonstrates a much better performance as demonstrated in \cref{tab:overall} and \cref{forget}. In L2P and DualPrompt, the prompts are originally designed for classification tasks, a single prompt could be selected by multiple tasks and trained offline in task-based scenarios, where forgetting could be greatly alleviated by using task identity and learning data repeatedly for each task. 
To solve F2OCL using L2P and DualPrompt, we remove all tricks depending on task identities, such as the label trick used to separate the parameters of linear classifier task-by-task, a prompt mask used in DualPrompt to select the corresponding prompt for each specific task.

Different from L2P and DualPrompt, we design our prompts for individual classes. The isolated prompts reduce the chance of forgetting due to shared prompts. Also, to learn as much as possible from data processed in one pass, we introduce contrastive prompt learning which explores information from current batch samples thoroughly but also considers the information from historical batches. To verify whether this is caused by the different classifiers used, we implemented L2P and DualPrompt with the same NCM classifier as our method to replace the linear classifier. The performances are even worse. So, we conclude the prompts designed in L2P and DualPrompt are unsuitable for F2OCL where training is based on processing data in one pass. 

In \cref{tab:overall} and \cref{forget}, our method demonstrates better performance against ViT-NCM. Since their difference is that our method uses prompts but ViT-NCM does not, it verifies the effectiveness of the proposed prompt for F2OCL.

\begin{figure}[t]
  \centering
   \includegraphics[width=0.9\linewidth]{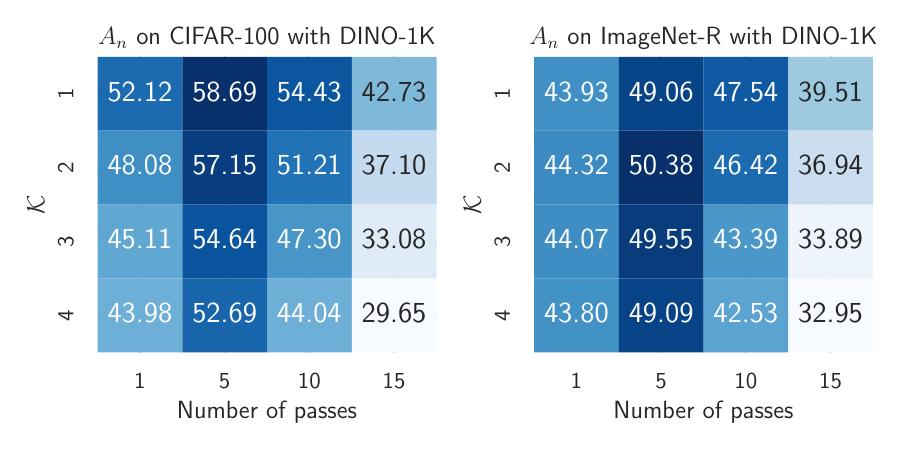}
   \vspace{-0.5cm}
   \caption{The impact to our method when processing data by multiple passes and selecting $\mathcal{K}$ keys from the prompt pool. }
   \label{heatmap}
   \vspace{-0.5cm}
\end{figure}

\begin{figure}[t]
  \centering    
   \includegraphics[width=0.9\linewidth]{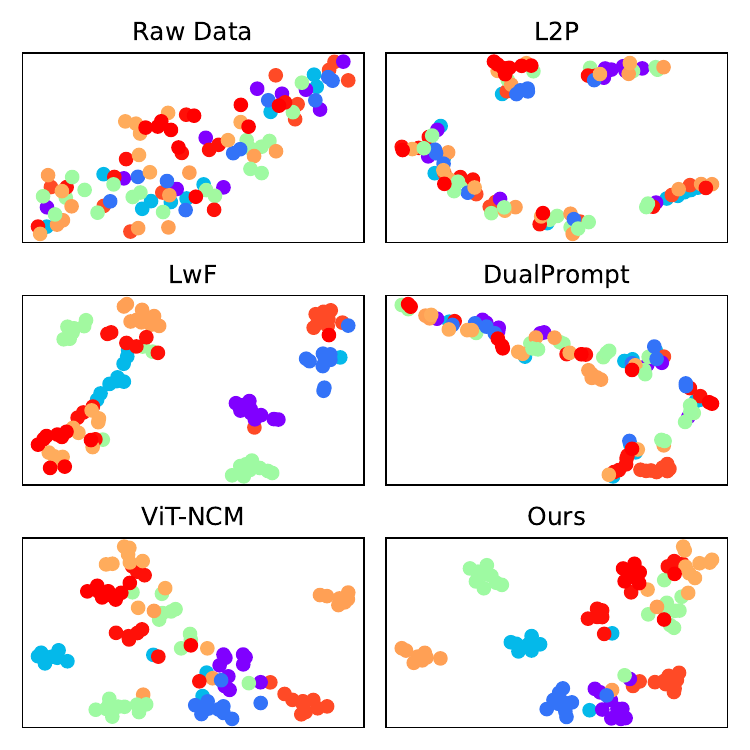}
   \caption{The embeddings distribution of 10 samples of 10 classes (CIFAR-100).}
   \label{distri}
   \vspace{-0.5cm}
\end{figure}

\subsection{Embedding Distribution}
In \cref{distri}, we visualize the distribution of 10 randomly selected samples of 10 randomly selected classes from the test set of CIFRA-100. The raw data of these samples are well mixed. For ViT-NCM, the embeddings of these samples, generated using ViT, are still mixed. For our method, the augmented embeddings of these samples are well separated. For other methods, each sample is represented by the probabilities estimated by the corresponding linear classifier to be different classes. The results in \cref{distri} show our method can disperse samples of different classes in a better way to support accurate classification.  

\subsection{Multiple Passes and Keys} \label{sec:further}
Since data is processed in one pass, it is interesting to know if the data in a batch can be processed multiple times before processing the next batch. Also, given a sample $\bm{x}$, it is possible to select $\mathcal{K}$ most similar keys of $\bm{q}_{\bm{x}}$ from the prompt pool as discussed in the paragraph following \cref{eq:11}. We test the effect of the two factors on the performance of our method and the results are reported in \cref{heatmap}. The best performances are achieved when the number of passes is $5$. The prompts are under-fitting when the number of passes is $1$. The prompts are over-fitting when the number of passes is $10$ and even worse when $15$. 

When the number of keys is 1, the best performance is achieved on CIFAR-100 for different numbers of passes. The situation is similar on ImageNet-R except that, when the number of passes is 5, the best performance is achieved for 2 keys. In general, more prompts lead to a performance decline. It implies more prompts tend to introduce more noise than useful information to the augmented embeddings of samples.   

\section{Conclusion}
Considering the privacy concerns of storing samples and the difficulty of identifying task boundaries in training data processed in one pass, this study proposes to investigate task-free and rehearsal-free online continual learning (F2OCL). The proposed solution integrates prompt learning with the NCM classifier. The prompt learning successfully addresses catastrophic forgetting without storing exemplars while the NCM classifier can continuously handle data without organizing them into learning tasks. This study deliberately designs an algorithm to learn prompts so that the performance of the NCM classifier is optimized. The effectiveness has been demonstrated by extensive experiments. 

\section{Acknowledgements} 
This work was supported by the Australian Research Council (ARC) Discovery Project Grant [DP210100743]. The authors gratefully acknowledge the financial support provided by the ARC.

\bibliographystyle{elsarticle-num} 
\bibliography{cas-refs}






\end{document}